\documentclass[10pt,twocolumn,letterpaper]{article}

\usepackage[pagenumbers]{wacv}

\usepackage{graphicx}
\usepackage{amsmath}
\usepackage{amssymb}
\usepackage{booktabs}
\usepackage[normalem]{ulem}
\useunder{\uline}{\ul}{}
\usepackage{multirow}
\usepackage{subcaption}

\usepackage[pagebackref,breaklinks,colorlinks]{hyperref}

\usepackage[capitalize]{cleveref}
\crefname{section}{Sec.}{Secs.}
\Crefname{section}{Section}{Sections}
\Crefname{table}{Table}{Tables}
\crefname{table}{Tab.}{Tabs.}

\def\wacvPaperID{1357} 
\def\confName{WACV}
\def\confYear{2025}

\begin{document}

\title{Representation Learning and Identity Adversarial Training for Facial Behavior Understanding}

\author{Mang Ning\\
Utrecht University\\
{\tt\small m.ning@uu.nl}
\and
Albert Ali Salah\\
Utrecht University\\
{\tt\small a.a.salah@uu.nl}
\and
Itir Onal Ertugrul\\
Utrecht University\\
{\tt\small i.onalertugrul@uu.nl}
}
\maketitle

\begin{abstract}

Facial Action Unit (AU) detection has gained significant attention as it enables the breakdown of complex facial expressions into individual muscle movements.  
In this paper, we revisit two fundamental factors in AU detection: diverse and large-scale data and subject identity regularization. 
Motivated by recent advances in foundation models, we highlight the importance of data and introduce Face9M, a diverse dataset comprising 9 million facial images from multiple public sources. 
Pretraining a masked autoencoder on Face9M yields strong performance in AU detection and facial expression tasks. More importantly, we emphasize that the Identity Adversarial Training (IAT) has not been well explored in AU tasks. To fill this gap, we first show that subject identity in AU datasets creates shortcut learning for the model and leads to sub-optimal solutions to AU predictions. Secondly, we demonstrate that strong IAT regularization is necessary to learn identity-invariant features. Finally, we elucidate the design space of IAT and empirically show that IAT circumvents the identity-based shortcut learning and results in a better solution. Our proposed methods, Facial Masked Autoencoder (FMAE) and IAT, are simple, generic and effective. Remarkably, the proposed FMAE-IAT approach achieves new state-of-the-art F1 scores on BP4D (67.1\%), BP4D+ (66.8\%), and DISFA (70.1\%) databases, significantly outperforming previous work. We release the code and model at \url{https://github.com/forever208/FMAE-IAT}.

\end{abstract}

\section{Introduction}
\label{sec:intro}

The Facial Action Coding System (FACS) was developed to objectively encode facial behavior through specific movements of facial muscles, named Action Units (AU) \cite{ekman1997face}. Compared with facial expression recognition (FER) \cite{lee2023latent, zhang2022learn, kim2022emotion} and valence and arousal estimation \cite{zhang2020m, meng2022valence, ning2023automated}, detecting action units offers a more nuanced and detailed understanding of human facial behavior capturing multiple individual facial actions simultaneously.

This problem attracted considerable interest within the deep learning community~\cite{chang2022knowledge, jacob2021facial, yang2018facial, onal2019d, she2021dive, yang2021exploiting}. Many works used a facial region prior \cite{li2017eac, onal2019d, chang2022knowledge}, introduced extra modalities \cite{zhang2023weakly, wang2024multi, zhang2024multimodal}, or incorporated the inherent AU relationships \cite{li2019semantic, luo2022learning, yang2023fan} to solve the AU detection task and achieved significant advancements. Diverging from these approaches, which often necessitate complex model designs or depend heavily on prior AU knowledge, in this paper, we revisit two fundamental factors that significantly contribute to the AU detection task: diverse and large-scale data and subject identity regularization.

Recently, data has become pivotal in training foundation models \cite{radford2021learning, alayrac2022flamingo, liu2024visual, team2024chameleon} and large language models~\cite{brown2020language, achiam2023gpt, touvron2023llama, chowdhery2023palm}. Following this trend, we introduce Face9M, a large-scale and diverse facial dataset curated and refined from publicly available datasets for pretraining. Different from contrastive learning methods \cite{bulat2022pre, chang2022knowledge, gao2024self},  we propose to do facial representation learning using Masked Autoencoders (MAE)~\cite{he2022masked}. The underlying motivation is that most  facial tasks require a fine-grained understanding of the face, and masked pretraining results in lower-level semantics than contrastive learning according to ~\cite{assran2023self}. Our large-scale facial representation learning approach demonstrates excellent generalization and scalability in downstream tasks. Notably, our proposed Facial Masked Autoencoder (FMAE), pretrained on Face9M, sets new state-of-the-art benchmarks in both AU detection and FER tasks.
    
Similar to the importance of data, domain knowledge and task-prior knowledge can be incorporated into the model in the form of regularization~\cite{gulrajani2017improved, ioffe2015batch, miyato2018spectral, ning2023input} to improve task performance. Our key observation is that popular AU detection benchmarks (i.e. BP4D \cite{zhang2014bp4d}, BP4D+ \cite{zhang2016multimodal}, DISFA \cite{mavadati2013disfa}) include at most 140 human subjects and 192,000 images, meaning that each subject has hundreds of annotated images. This abundance can lead models to prefer simple, easily recognizable patterns over more complex but generalizable ones, as suggested by the \textbf{shortcut learning theory} \cite{geirhos2020shortcut, hermann2023foundations}. Therefore, we hypothesize that \textbf{AU detection models tend to learn the subject identity
features to infer the AUs, resulting in learning a trivial solution that does not generalize well}. To verify our hypothesis, we employed the linear probing technique — adding a learnable linear layer to a trained AU model while freezing the network backbone —to measure identity recognition accuracy. The high accuracy (83\%) we obtained in predicting the identities of the subjects clearly shows that the models effectively `memorize' subject identities. To counteract the learning of identity-based features, we propose in this paper Identity Adversarial Training (IAT) for AU detection task by adding a linear identity prediction head and unlearning the identity feature using gradient reverse~\cite{ganin2015unsupervised}. Further analysis shows that IAT significantly reduces the identity accuracy of linear probing and leads to better learning dynamics that avoid convergence to trivial solutions. This method further improves our AU models beyond the advantages brought by pretraining with a large-scale dataset.

Although Zhang et al. \cite{zhang2018identity} first introduced identity-based adversarial training to AU detection tasks, the identity learning issue and its negative effect (identity shortcut learning) have not been explored. Also, the design space of IAT lacks illustration in \cite{zhang2018identity}. We revisit the identity adversarial training method in depth to answer these unexplored questions. In contrast to the weak identity regularization used in \cite{zhang2018identity}, we demonstrate that AU detection requires a strong identity regularization. To this end, the linear identity head and a large gradient reverse scaler are necessities for the AU detection task. Our proposed FMAE with IAT sets a new record of F1 score on BP4D (67.1$\%$), BP4D+ (66.8$\%$) and DISFA (70.1$\%$) datasets, substantially surpassing previous work. 

    Overall, the main contributions of this paper are:

\begin{itemize}
    \item We demonstrate the effectiveness of using a diverse dataset for facial representation learning.
    \item We highlight the identity shortcut learning issue and propose the use of a linear identity head and a large gradient reversal scalar in IAT to mitigate this issue for AU detection.
    \item We release the code and checkpoint of FMAE with various model sizes (small, base, large), aiming at facilitating all facial tasks.
\end{itemize}

\section{Related Work}
\label{sec: related work}

\subsection{Action Unit Detection}
Recent works have proposed several deep learning-based approaches for facial action unit (AU) detection. Some of them have divided the face into multiple regions or patches \cite{zhao2016deep, li2017eac, onal2019d} to learn AU-specific representations and some have explicitly modeled the relationships among AUs \cite{li2019semantic, luo2022learning, yang2023fan}. The most recent approaches have focused on detecting AUs using vision transformers on RGB images \cite{jacob2021facial} and on multimodal data including RGB, depth images, and thermal images \cite{zhang2024multimodal}. Yin et al. \cite{yin2024fg} have used generative models to extract representations and a pyramid CNN interpreter to detect AUs. Yang et al. \cite{yang2023toward} jointly modeled AU-centered features, AU co-occurrences, and AU dynamics. Contrastive learning has recently been adopted for AU detection \cite{li2023contrastive, shang2024learning}. Particularly, Chang et al. \cite{chang2022knowledge} have adopted contrastive learning among AU-related regions and performed predictive training considering the relationships among AUs. Zhang et al. \cite{zhang2023weakly} have proposed a weakly-supervised text-driven contrastive approach using coarse-grained activity information to enhance feature representations. In addition to fully supervised approaches, Tang et al. \cite{tang2021piap} have implemented a semi-supervised approach with discrete feedback. However, none of these approaches have made use of large-scale self-supervised pretraining.

\subsection{Facial Representation Learning}
Facial representation learning~\cite{bulat2022pre, cai2023marlin, zheng2022general} has seen substantial progress with the advent of self-supervised learning~\cite{chen2020simple, he2020momentum, bao2021beit, he2022masked, caron2021emerging}. For example, Mask Contrastive Face~\cite{wang2023toward} combines mask image modeling with contrastive learning to do self-distillation, thereby enhancing facial representation quality. Similarly, ViC-MAE~\cite{hernandez2023vic} integrates MAE with temporal contrastive learning to enhance video and image representations. MAE-face~\cite{ma2024facial} uses MAE for facial pertaining by 2 million facial images. Additionally, ContraWarping~\cite{xue2023unsupervised} employs global transformations and local warping to generate positive and negative samples for facial representation learning. To learn good local facial representations, Gao et al. \cite{gao2024self} explicitly enforce the consistency of facial regions by matching the local facial representations across views. Different from the above-mentioned work that mainly focuses on models, we emphasize the importance of data (diversity and quantity). Our collected datasets contain 9 million images from various public resources.

\subsection{Adversarial Training and Gradient Reverse}
Adversarial training \cite{goodfellow2014explaining} is a regularization technique in deep learning to enhance the model's robustness specifically against input perturbations that could lead to incorrect outputs. Although gradient reverse technique \cite{ganin2015unsupervised} aims to minimize domain discrepancy for better generalization across different data distributions, these two techniques share the same spirit of the 'Min-Max' training paradigm and are used to improve the model robustness \cite{kurakin2016adversarial, madry2017towards, ganin2016domain, tzeng2017adversarial}. Gradient reverse has also been used for the regularization of fairness \cite{raff2018gradient} or for meta-learning \cite{andrychowicz2016learning}. 

The most relevant research to our paper is \cite{zhang2018identity}, where the authors introduce identity-based adversarial training for the AU detection task. However, they did not thoroughly investigate the identity learning phenomenon and its detrimental impacts. Moreover, their empirical settings, the small gradient reverse scaler and the 2-layer MLP identity head, have been   \cite{zhang2018identity} verified as an inferior solution to AU detection. By contrast, we conduct a comprehensive examination for IAT to address these unexplored questions.

\section{Methods}
\label{sec: methods}

\subsection{Large-scale Facial MAE Pretraining}
While the machine learning community has long established the importance of having rich and diverse data for training, recent successes in foundation models and large language models illustrated the full potential of pretaining \cite{radford2021learning, liu2024visual, brown2020language, achiam2023gpt, touvron2023llama}. In line with this, our research pivots towards a nuanced exploration of data diversity and quantification in the context of facial representation learning. Unlike natural image datasets like ImageNet-1k, face datasets have low variance. Also, we observe that different facial datasets have domain shifts regarding the facial area, perspective and background. To increase the data diversity, we propose to collect a large facial dataset for pertaining from multiple data sources.



We first collect facial images from CelebA ~\cite{liu2018large}, FFHQ ~\cite{karras2019style}, VGGFace2 \cite{cao2018vggface2}, CASIA-WebFace \cite{yi2014learning}, MegaFace \cite{kemelmacher2016megaface}, EDFace-Celeb-1M \cite{zhang2022edface}, UMDFaces \cite{bansal2017umdfaces} and LAION-Face \cite{zheng2022general} datasets, because these datasets contain a massive number of identities collected in diverse scenarios. For instance, the facial images in UMDFaces also capture the upper body with various image sizes, while some datasets (FFHQ, CASIA-WebFace) mainly feature the center face. We then discard images whose width-height-ratio or height-width-ratio is larger than 1.5. Finally, the remaining images are resized to 224*224. The whole process yields 9 million facial images (termed Face9M) which will be used for self-supervised facial pertaining.

Regarding representation learning methods, we apply Masked Image Modeling (MIM) \cite{he2022masked} as it tends to learn more fine-grained features than contrastive learning according to the study in \cite{assran2023self}, which benefits facial behavior understanding. Specifically, we utilize Face9M to train a masked autoencoder (MAE) by the mean squared error between the reconstructed and original images in the pixel space. The resulting model is termed FMAE, and the decoder of FMAE is discarded for the downstream tasks.

\subsection{Identity Adversarial Training}
\label{sec: IAT method}
One of our key findings in this paper is that, the limited number of subjects in AU datasets makes identity recognition a trivial task and provides a shortcut learning path, resulting in a AU model that contains identity-related features and does not generalize well (see Section \ref{sec: IAT}). Motivated by the gradient reverse in domain adaption \cite{ganin2015unsupervised}, we propose to apply gradient reverse on AU detection to learn identity-invariant features, aiming at better model generalization.

\begin{figure}[htbp]
\begin{center}
\centerline{\includegraphics[width=\columnwidth]{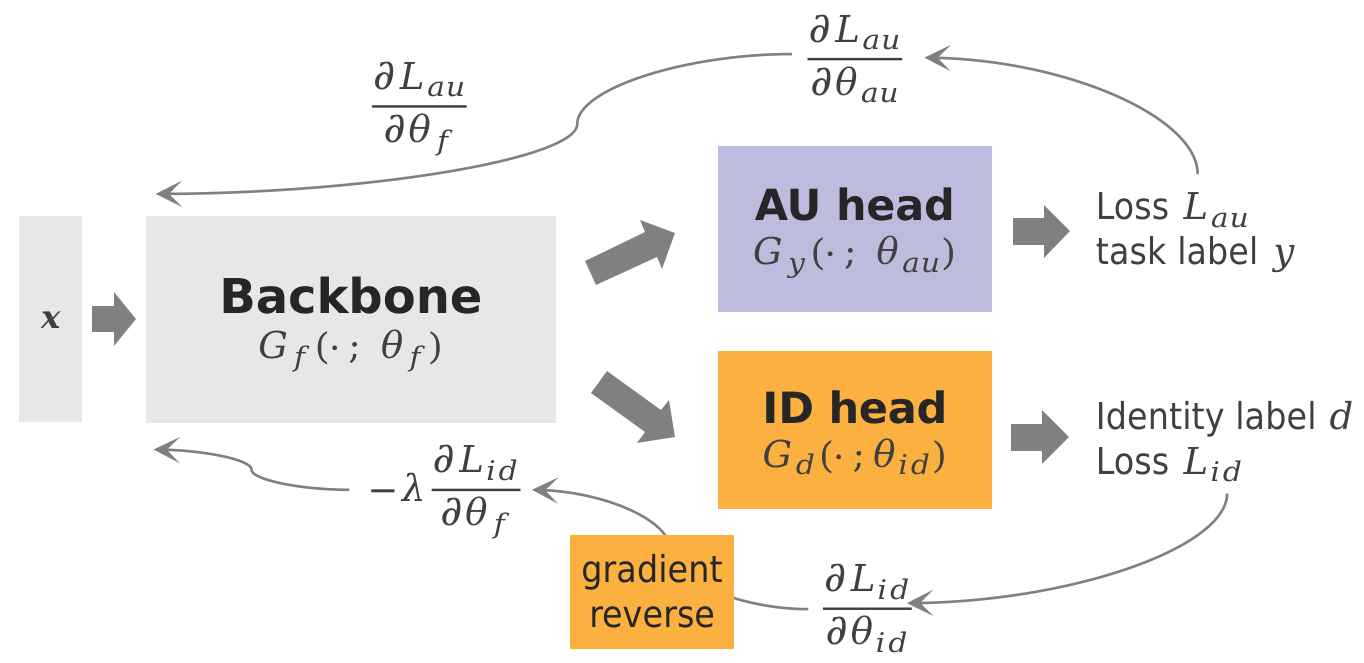}}
\caption{Architecture of Identity Adversarial Training. The AU head and ID head both are a linear classifier predicting the AUs and identity, respectively. The backbone $G_{f}(\cdot; \theta_{f})$ is the encoder of the pretrained FMAE. During training, the AU head is optimized by $\frac{\partial L_{au}}{\partial \theta_{f}}$ and the ID head is optimized by $\frac{\partial L_{id}}{\partial \theta_{f}}$. The gradient reverse layer multiplies the gradient by a negative value $-\lambda$ to unlearn the features capable of recognizing identities. Finally, the parameters of the backbone are optimized by the two forces: $-\lambda \frac{\partial L_{id}}{\partial \theta_{f}}$ and $\frac{\partial L_{au}}{\partial \theta_{f}}$.
}
\label{fig: architecture}
\end{center}
\vskip -0.1in
\end{figure}

Our model architecture is presented in Figure \ref{fig: architecture}, where the backbone is a vision transformer and parameterized by $\theta_{f}$, the AU head predicts the AUs and the ID head outputs the subject identities, respectively. The input image $\pmb{x}$ is first mapped by the backbone $G_{f}(\cdot; \theta_{f})$ to a D-dimensional feature vector $\pmb{f} \in \mathbb{R}^{D}$, then the feature vector $\pmb{f}$ is fed into the AU head $G_{f}(\cdot; \theta_{au})$ and the ID head $G_{f}(\cdot; \theta_{id})$. simultaneously. Assume that we have data samples $(\pmb{x}, y, d) \sim D_{s}$, parameters $\theta_{au}$ of the AU head are optimized to minimize the AU loss $L_{au}$ given AU label $y$, and parameters $\theta_{id}$ of the ID head are trained to minimize the identity loss $L_{id}$ given the identity label $d$.

To make the feature vector $\pmb{f}$ invariant to subject identity, we seek the parameters $\theta_{f}$ of the backbone that 
\textbf{maximize} the identity loss $L_{id}$ (Equation \ref{eq: solution1}), so that the backbone excludes the identity-based features. In the meantime, the backbone $G_{f}(\cdot; \theta_{f})$ is expected to minimize the AU loss $L_{au}$. Formally, we consider the following functional loss:

\begin{equation}
\label{eq: loss_au}
L_{au} = \mathbb{E}_{(\pmb{x}, y) \sim D_{s}}[CE(G_{y}(G_{f}(\pmb{x}; \theta_{f}); \theta_{au}), \ y)]
\end{equation}

\noindent
\begin{equation}
\label{eq: loss}
L_{id} = \mathbb{E}_{(\pmb{x}, d) \sim D_{s}}[CE(G_{d}(G_{f}(\pmb{x}; \theta_{f}); \theta_{id}), \ d)]
\end{equation}
where $CE$ denotes the cross entropy loss function. We seek the parameters $\theta_{f}^{*}$, $\theta_{au}^{*}$, $\theta_{id}^{*}$ that deliver a solution:

\noindent
\begin{equation}
\label{eq: solution1}
(\theta_{f}^{*}, \theta_{au}^{*}) = arg \underset{\theta_{f}, \theta_{au}}{min}  L_{au}(D_s; \theta_{f}, \theta_{au}) - \lambda L_{id}(D_s; \theta_{f}, \theta_{id}^{*})
\end{equation}

\noindent
\begin{equation}
\label{eq: solution2}
\theta_{id}^{*} = arg \ \underset{\theta_{id}}{min} \ L_{id}(D_s; \theta_{f}^{*}, \theta_{id})
\end{equation}

\noindent
where the parameter $\lambda$ controls the trade-off between the two objectives that shape the feature $\pmb{f}$ during learning. Comparing the identity loss $L_{id}$ in Equation \ref{eq: solution1} and Equation \ref{eq: solution2}, $\theta_{f}$ is optimized to maximize to increase $L_{id}$ while $\theta_{id}$ is learned to reduce $L_{id}$. To achieve these two opposite optimizations through regular gradient descent and backpropagation, the gradient reverse layer is designed to reverse the identity partial derivative $\frac{\partial L_{id}}{\partial \theta_{id}}$ before it is propagated to the backbone. The resultant derivative $-\lambda \frac{\partial L_{id}}{\partial \theta_{f}}$, together with $\frac{\partial L_{au}}{\partial \theta_{f}}$, are used to update the backbone parameter $\theta_{f}$. 

Intuitively, the backbone is still optimized to learn the AU-related features, but under the force of reducing the identity-related features. The `Min-Max' training paradigm in gradient reverse (see Equation \ref{eq: solution1})  resembles the adversarial training~\cite{madry2018towards} and Generative Adversarial Networks (GANs)~\cite{goodfellow2020generative}, so we name our method `Identity Adversarial Training' for the AU detection task.

Importantly, we reveal the key design of identity adversarial training for AU detection: \textbf{a strong adversarial regularization (large magnitude of $-\lambda \frac{\partial L_{id}}{\partial \theta_{f}}$) is required to learn identity-invariant features for the backbone}. Specifically, we propose to use a large $\lambda$ and a linear projection layer for the ID head. The former scales up the $\frac{\partial L_{id}}{\partial \theta_{f}}$ and the latter ensures a large $L_{id}$, leading to a large  $||-\lambda \frac{\partial L_{id}}{\partial \theta_{f}}||$ during training. In Section \ref{sec: strong adversarial}, we will show that the small $\lambda$ and 2-layer MLP ID head used by \cite{zhang2018identity} would lead to a weak identity regularization (small magnitude of $-\lambda \frac{\partial L_{id}}{\partial \theta_{f}}$) and inferior AU performance. We defer more details and analysis to Section \ref{sec: strong adversarial}

\begin{table*}[tb]
\centering
\small
\caption{F1 scores (in \%) achieved for 12 AUs on BP4D dataset. The best and the second-best results of each column are indicated with
bold font and underline, respectively.}
\begin{tabular}{@{}lllllllllllllll@{}}
\toprule
\multirow{2}{*}{Methods} & \multirow{2}{*}{Venue} & \multicolumn{12}{c}{AU} & \multirow{2}{*}{\textbf{Avg.}} \\ \cmidrule(lr){3-14}
 &  & 1 & 2 & 4 & 6 & 7 & 10 & 12 & 14 & 15 & 17 & 23 & 24 &  \\ \midrule
HMP-PS~\cite{song2021hybrid} & \multicolumn{1}{l|}{CVPR'21} & 53.1 & 46.1 & 56.0 & 76.5 & 76.9 & 82.1 & 86.4 & 64.8 & 51.5 & 63.0 & 49.9 & 54.5 & 63.4 \\
SEV-Net~\cite{yang2021exploiting} & \multicolumn{1}{l|}{CVPR'21} & 58.2 & 50.4 & 58.3 & \textbf{81.9} & 73.9 & \textbf{87.8} & 87.5 & 61.6 & 52.6 & 62.2 & 44.6 & 47.6 & 63.9 \\
FAUT~\cite{jacob2021facial} & \multicolumn{1}{l|}{CVPR'21} & 51.7 & 49.3 & 61.0 & 77.8 & 79.5 & 82.9 & 86.3 & 67.6 & 51.9 & 63.0 & 43.7 & 56.3 & 64.2 \\
PIAP~\cite{tang2021piap} & \multicolumn{1}{l|}{ICCV'21} & 55.0 & 50.3 & 51.2 & {\ul 80.0} & 79.7 & 84.7 & \textbf{90.1} & 65.6 & 51.4 & 63.8 & 50.5 & 50.9 & 64.4 \\
KSRL~\cite{chang2022knowledge} & \multicolumn{1}{l|}{CVPR'22} & 53.3 & 47.4 & 56.2 & 79.4 & \textbf{80.7} & 85.1 & 89.0 & 67.4 & 55.9 & 61.9 & 48.5 & 49.0 & 64.5 \\
ANFL~\cite{luo2022learning} & \multicolumn{1}{l|}{IJCAI'22} & 52.7 & 44.3 & 60.9 & 79.9 & 80.1 & {\ul85.3} & 89.2 & {\ul 69.4} & 55.4 & 64.4 & 49.8 & 55.1 & 65.5 \\
CLEF~\cite{zhang2023weakly} & \multicolumn{1}{l|}{ICCV'23} & 55.8 & 46.8 & \textbf{63.3} & 79.5 & 77.6 & 83.6 & 87.8 & 67.3 & 55.2 & 63.5 & 53.0 & 57.8 & 65.9 \\
MCM~\cite{zhang2024multimodal} & \multicolumn{1}{l|}{WACV'24} & 54.4 & 48.5 & 60.6 & 79.1 & 77.0 & 84.0 & 89.1 & 61.7 & \textbf{59.3} & 64.7 & 53.0 & \textbf{60.5} & 66.0 \\
AUFormer~\cite{yuan2024auformer} & \multicolumn{1}{l|}{ECCV'24} & - & - & - & - & - & - & - & - & - & - & - & - & 66.2 \\ 
MDHR~\cite{wang2024multi} & \multicolumn{1}{l|}{CVPR'24} & 58.3 & {\ul 50.9} & 58.9 & 78.4 & {\ul80.3} & 84.9 & 88.2 & \textbf{69.5} & {\ul 56.0} & \textbf{65.5} & 49.5 & {\ul59.3} & {\ul 66.6} \\ \midrule
\textbf{FMAE} & \multicolumn{1}{l|}{(ours)} & {\ul59.2} & 50.0 & {\ul62.7} & 
{\ul80.0} & 79.2 & 84.7 & 89.8 & 63.5 & 52.8 & 65.1 & 
\textbf{55.3} & 56.9 & {\ul 66.6} \\
\textbf{FMAE-IAT} & \multicolumn{1}{l|}{(ours)} & \textbf{62.7} & \textbf{51.9} & {\ul62.7} & 79.8 & 80.1 & 84.8 & {\ul89.9} & 64.6 & 54.9 & {\ul65.4} & 
{\ul53.1} & 54.7 & \textbf{67.1} \\ \bottomrule
\end{tabular}
\label{tab: BP4D results}
\end{table*}

\section{Experiments}
\label{sec: experiments}
We test the performance of FMAE and FMAE-IAT on AU benchmarks, using the F1 score. To illustrate the representation learning efficacy of FMAE, we also report its facial expression recognition (FER) accuracy on RAF-DB \cite{shan2018reliable} and AffectNet \cite{mollahosseini2017affectnet} databases, and compare FMAE with previous face models pretrained based on contrastive learning.

\subsection{Datasets}
\textbf{BP4D} \cite{zhang2014bp4d} is a manually annotated database of spontaneous behavior containing videos of 41 subjects. There are 8 activities designed to elicit various spontaneous emotional responses, resulting in 328 video clips. A total of 140,000 frames are annotated by expert FACS annotators. Following~\cite{li2017eac, zhang2023weakly, wang2024multi}, we split all annotated frames into three subject-exclusive folds for 12 AUs.

\textbf{BP4D+} \cite{zhang2016multimodal} is an extended dataset of BP4D and features 140 participants. For each subject, 20 seconds from 4 activities are manually annotated by FACS annotators, resulting in 192,000 labelled frames. We divide the subjects into four folds as per guidelines in~\cite{zhang2021multi, zhang2023weakly} and 12 AUs are used for AU detection.

\textbf{DISFA} \cite{mavadati2013disfa} contains left-view and right-view videos of 27 subjects. Similar to~\cite{yang2021exploiting, zhang2023weakly}, we use 8 of 12 AUs. We treat samples with AU intensities higher or equal to 2 as positive samples. The database contains 130,000 manually annotated images. Following \cite{zhang2023weakly} we perform subject-exclusive 3-fold cross-validation.

\textbf{RAF-DB} \cite{shan2018reliable} contains 15,000 facial images with annotations for 7 basic expressions namely neutral, happiness, surprise, sadness, anger, disgust, and fear. Following the previous work \cite{she2021dive, zhang2023weakly}, we use 12,271 images for training and the remaining 3,068 for testing.

\textbf{AffectNet} \cite{mollahosseini2017affectnet} is currently the largest FER dataset with annotations for 8 expressions (neutral, happy, angry, sad, fear, surprise, disgust, contempt). AffectNet-8 includes all expression images with 287,568 training samples
and 4,000 testing samples. In practice, we only use 37,553 images (from Kaggle) for training as training on the whole training set is expensive.

\subsection{Implementation details}
Regarding facial representation learning, we pretrain FMAE with Face9M for 50 epochs (including two warmup epochs) using four NVIDIA A100 GPUs. The remaining parameter settings follow \cite{he2022masked} without any changes. After the pertaining, we finetune FMAE for FER tasks with cross-entropy loss, and fine-tune FMAE and FMAE-IAT for AU detection with binary cross-entropy loss. In most cases, we finetune the model for 30 epochs with a batch size of 64 and a base learning rate of 0.0005. Following MAE \cite{he2020momentum}, we use a weight decay of 0.05, AutoAugmentation \cite{cubuk2019autoaugment} and Random Erasing 0.25 \cite{zhong2020random} for regularization. By default, we apply ViT-large for FMAE and FMAE-IAT throughout this paper, if not specified otherwise. The complete code, hyperparameters and training/testing protocols are posted on our GitHub repository for reproducibility.   

\begin{figure}[htb]
\begin{center}
\centerline{\includegraphics[width=0.9\columnwidth]{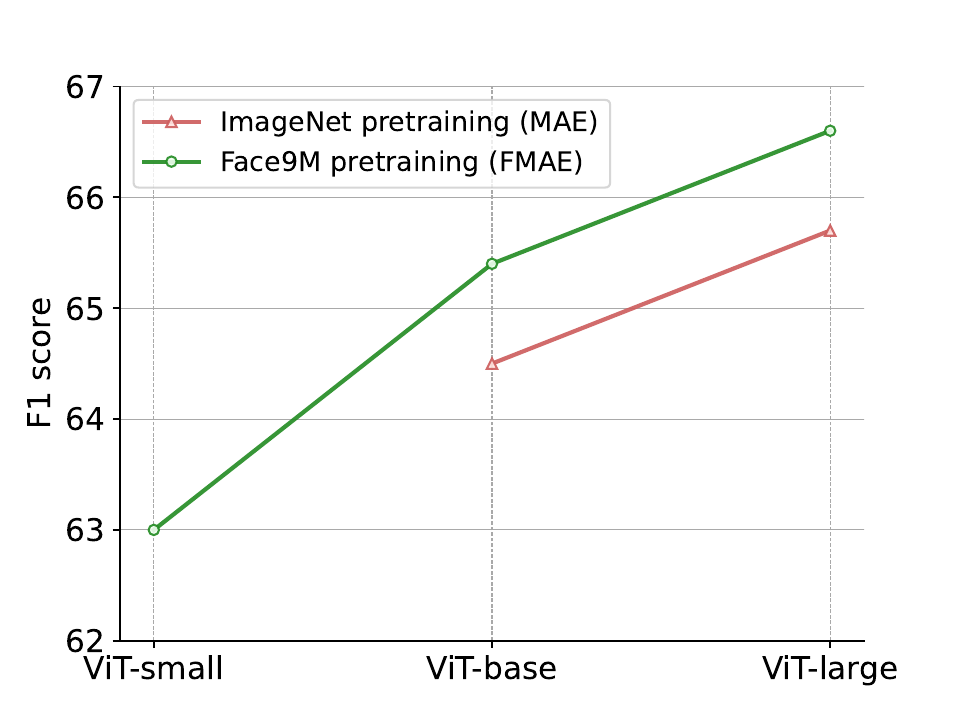}}
\caption{F1 results of FMAE using different model sizes on 12 AUs of the BP4D. Models pretrained on Face9M are better than the ones pretrained on ImageNet-1k. MAE paper does not train ViT-small on ImageNet-1k, thus this entry is missing.}
\label{fig: MAE scale}
\end{center}
\vskip -0.3in
\end{figure}

\subsection{Result of FMAE}
We first show the F1 score of FMAE on the BP4D dataset in Table \ref{tab: BP4D results}. FMAE achieves the same average F1 (66.6\%) with the state-of-the-art method MDHR \cite{wang2024multi} which utilizes a two-stage model to learn the hierarchical AU relationships. Here, we see the effectiveness of data-centric facial representation learning, and demonstrate that a simple vision transformer \cite{dosovitskiy2020image}, which is the architecture of FMAE, is capable of learning complex AU relationships. FMAE surpasses all previous work on BP4D+ and DISFA  by achieving 66.2\% and 68.7\% F1 scores, respectively (see Table \ref{tab: BP4D plus results} and Table \ref{tab: DISFA results}).

\begin{table*}[tb]
\centering
\small
\caption{F1 scores (in \%) achieved for 12 AUs on BP4D+ dataset. The best and the second-best results of each column are indicated with
bold font and underline, respectively. MFT* uses extra depth modality.}
\begin{tabular}{@{}lllllllllllllll@{}}
\toprule
\multirow{2}{*}{Methods} & \multirow{2}{*}{Venue} & \multicolumn{12}{c}{AU} & \multirow{2}{*}{\textbf{Avg.}} \\ \cmidrule(lr){3-14}
 &  & 1 & 2 & 4 & 6 & 7 & 10 & 12 & 14 & 15 & 17 & 23 & 24 &  \\ \midrule
ViT~\cite{dosovitskiy2020image} & \multicolumn{1}{l|}{ICLR'21} & 45.6 & 38.2 & 35.5 & 85.9 & 88.3 & 90.3 & 89.0 & 81.9 & 45.8 & 48.8 & 57.2 & 34.6 & 61.6 \\
CLIP~\cite{radford2021learning} & \multicolumn{1}{l|}{ICML'21} & 49.4 & 39.7 & 38.9 & 85.7 & 87.6 & 90.6 & 89.0 & 80.6 & 44.9 & 50.3 & 56.1 & 32.8 & 62.1 \\
SEV-Net~\cite{yang2021exploiting} & \multicolumn{1}{l|}{CVPR'21} & 47.9 & 40.8 & 31.2 & \textbf{86.9} & 87.5 & 89.7 & 88.9 & \textbf{82.6} & 39.9 & {\ul 55.6} & 59.4 & 27.1 & 61.5 \\
MFT~\cite{zhang2021multi} & \multicolumn{1}{l|}{FG'21} & 48.4 & 37.1 & 34.4 & 85.6 & {\ul 88.6} & 90.7 & 88.8 & 81.0 & 47.6 & 51.5 & 55.6 & 36.9 & 62.2 \\
MFT*~\cite{zhang2021multi} & \multicolumn{1}{l|}{FG'21} & 49.6 & 42.0 & 43.5 & 85.8 & {\ul 88.6} & 90.6 & 89.7 & 80.8 & 49.8 & 52.2 & 59.1 & 38.4 & 64.2 \\
CLEF~\cite{zhang2023weakly} & \multicolumn{1}{l|}{ICCV'23} & 47.5 & 39.6 & 40.2 & 86.5 & 87.3 & 90.5 & \textbf{89.9} & 81.6 & 47.0 & 46.6 & 54.3 & 41.5 & 63.1 \\
GLTE-Net~\cite{an2024learning} & \multicolumn{1}{l|}{Intelli'24} & 51.5 & {\ul 46.6} & 43.5 & {\ul 86.8} & \textbf{89.6} & {\ul 91.0} & {\ul 89.8} & 82.3 & 46.8 & 49.3 & {\ul 60.9} & \textbf{50.9} & 65.7 \\ \midrule
\textbf{FMAE} & \multicolumn{1}{l|}{(ours)} & {\ul 53.9} & 45.5 & {\ul 45.9} & 86.2 & 88.3 & \textbf{91.2} & \textbf{89.9} & 82.3 & \textbf{51.3} & \textbf{56.3} & 60.7 & 42.7 & {\ul 66.2} \\
\textbf{FMAE-IAT} & \multicolumn{1}{l|}{(ours)} & \textbf{54.2} & \textbf{47.0} & \textbf{53.9} & 85.7 & 88.4 & \textbf{91.2} & 89.7 & {\ul 82.4} & {\ul 50.3} & 54.4 & \textbf{61.0} & {\ul 43.4} & \textbf{66.8} \\ \bottomrule
\end{tabular}
\label{tab: BP4D plus results}
\end{table*}

\begin{table*}[tb]
\centering
\small
\caption{F1 scores (in \%) achieved for 8 AUs on DISFA dataset. The best and the second-best results of each column are indicated with
bold font and underline, respectively. 
}
\begin{tabular}{@{}lllllllllll@{}}
\toprule
\multirow{2}{*}{Methods} & \multirow{2}{*}{Venue} & \multicolumn{8}{c}{AU} & \multirow{2}{*}{\textbf{Avg.}} \\ \cmidrule(lr){3-10}
 &  & 1 & 2 & 4 & 6 & 9 & 12 & 25 & 26 &  \\ \midrule
FAUT~\cite{jacob2021facial} & \multicolumn{1}{l|}{CVPR'21} & 46.1 & 48.6 & 72.8 & 56.7 & 50.0 & 72.1 & 90.8 & 55.4 & 61.5 \\
PIAP~\cite{tang2021piap} & \multicolumn{1}{l|}{ICCV'21} & 50.2 & 51.8 & 71.9 & 50.6 & 54.5 & {\ul 79.7} & 94.1 & 57.2 & 63.8 \\
ANFL~\cite{luo2022learning} & \multicolumn{1}{l|}{IJCAI'22} & 54.6 & 47.1 & 72.9 & 54.0 & 55.7 & 76.7 & 91.1 & 53.0 & 63.1 \\
KSRL~\cite{chang2022knowledge} & \multicolumn{1}{l|}{CVPR'22} & 60.4 & 59.2 & 67.5 & 52.7 & 51.5 & 76.1 & 91.3 & 57.7 & 64.5 \\
KS~\cite{li2023knowledge} & \multicolumn{1}{l|}{ICCV'23} & 53.8 & 59.9 & 69.2 & 54.2 & 50.8 & 75.8 & 92.2 & 46.8 & 62.8 \\
CLEF~\cite{zhang2023weakly} & \multicolumn{1}{l|}{ICCV'23} & 64.3 & {\ul 61.8} & 68.4 & 49.0 & 55.2 & 72.9 & 89.9 & 57.0 & 64.8 \\
SACL~\cite{liu2023multi} & \multicolumn{1}{l|}{TAC'23} & 62.0 & \textbf{65.7} & 74.5 & 53.2 & 43.1 & 76.9 & \textbf{95.6} & 53.1 & 65.5 \\
MDHR~\cite{wang2024multi} & \multicolumn{1}{l|}{CVPR'24} & \textbf{65.4} & 60.2 & {\ul 75.2} & 50.2 & 52.4 & 74.3 & 93.7 & 58.2 & 66.2 \\
AUFormer~\cite{yuan2024auformer} & \multicolumn{1}{l|}{ECCV'24} & - & - & - & - & - & - & - & - & 66.4 \\
GPT-4V~\cite{lu2024gpt} & \multicolumn{1}{l|}{CVPRW'24} & 52.6 & 56.4 & \textbf{82.9} & \textbf{64.3} & 55.3 & 75.4 & 91.2 & 66.4 & 67.3 \\ \midrule
\textbf{FMAE} & \multicolumn{1}{l|}{(ours)} & 62.7 & 59.5 & 67.3 & 55.6 & \textbf{61.8} & 77.9 & 95.0 & {\ul 69.8} & {\ul 68.7} \\
\textbf{FMAE-IAT} & \multicolumn{1}{l|}{(ours)} & {\ul 64.7} & 61.3 & 70.8 & {\ul 58.1} & {\ul 59.4} & \textbf{79.9} & {\ul 95.2} & \textbf{71.3} & \textbf{70.1} \\ \bottomrule
\end{tabular}
\label{tab: DISFA results}
\end{table*}

To further verify the importance of the Face9M dataset, we compare FMAE pretrained on Face9M with FMAE pretrained on ImageNet-1k \cite{deng2009imagenet}, using BP4D as the test set. Figure \ref{fig: MAE scale} shows that FMAE pretrained on Face9M always outperforms the one pretrained on ImageNet-1k given the same model size (ViT-base or ViT-large). Also, we empirically demonstrate that FMAE benefits from the scaling effect of model size on AU detection tasks (see the green line in Figure \ref{fig: MAE scale}).

\begin{table}[th]
\small
\centering
\caption{Results of accuracy on FER benchmarks. FMAE surpasses all previous contrastive-related work.}
\begin{tabular}{@{}lll|ll@{}}
\toprule
Model & Contrastive & MIM & AffectNet-8 & RAF-DB \\ \midrule
MCF \cite{wang2023toward} & $\surd$ & $\surd$ & 60.98 & 86.86 \\
FaRL \cite{zheng2022general} & $\surd$ & $\surd$ & - & 88.31 \\
CLEF \cite{zhang2023weakly} & $\surd$ &  & 62.77 & 90.09 \\
FRA \cite{gao2024self} & $\surd$ &  & - & 90.76 \\
LA-Net \cite{wu2023net} & $\surd$ &  & 64.54 & 91.78 \\ \midrule
FMAE (ours) &  & $\surd$ & \textbf{64.79} & \textbf{93.45} \\ \bottomrule
\end{tabular}

\label{tab: FER results}
\end{table}

In addition to AU detection, we benchmark FMAE on the downstream facial task of FER to verify the effectiveness of masked image representation learning. We present the results of FMAE on AffectNet-8 and RAF-DB in Table \ref{tab: FER results} and compare FMAE with other contrastive learning-based models.  FMAE sets a new state-of-art accuracy on both datasets (64.79\% on AffectNet-8 and 93.45\% on RAF-DB). Note that, we did not test FMAE-IAT on FER tasks because these datasets do not include the identity labels and do not suffer from identity shortcut learning due to the large number of subjects.

\subsection{Results of FMAE-IAT}
Although FMAE has already achieved superior results on AU benchmarks, we highlight that the Identity Adversarial Training could further boost the performance of FMAE across all AU datasets. Specifically, we compare FMAE-IAT with the most recent state of the art methods on BP4D, BP4D+ and DISFA datasets. Table \ref{tab: BP4D results} suggests that FMAE-IAT shows superior performance by achieving an average F1 Score of 67.1\% and FMAE-IAT ranks as the best or second-best performer in several individual AUs, notably AU 1, 2, 4, 12, 17 and 23. Similarly, FMAE-IAT also stands out on BP4D+ dataset with the highest average F1 score of 66.8\% shown in Table \ref{tab: BP4D plus results}. Our results on the DISFA benchmark given in Table \ref{tab: DISFA results} are even more distinguishing, FMAE-IAT gains the best or the second-best performance on 6 out of 8 AUs, pushing the average F1 score beyond the 70\% mark.

For the gradient reverse layer, we use $\lambda=2.0$ for BP4D, $\lambda=1.0$ for BP4D+ and $\lambda=0.5$ for DISFA. Differing from the setting $\lambda\in[0.008, 0.08]$ used in \cite{zhang2018identity}, we emphasize that a strong IAT regularization is necessary for AU tasks and we defer the in-depth discussion throughout Section \ref{sec: IAT}.

\section{Analysis of Identity Adversarial Training}
\label{sec: IAT}

In this section, we elucidate IAT by first showing the identity learning issue in AU tasks (Section \ref{sec: ID learning}), then demonstrating the learning dynamics refined by IAT (Section \ref{sec: learning dynamics}), and finally illustrating the importance of ensuring a strong regularization of IAT (Section \ref{sec: strong adversarial}).

\subsection{Linear probing for identity recognition}
\label{sec: ID learning}
Motivated by the shortcut learning theory \cite{geirhos2020shortcut, hermann2023foundations}, we hypothesize that each subject of AU datasets is exposed to the neural network hundreds of times in a single training epoch, which provides an identity shortcut for the model to learn the subject identity. This identity learning issue is undesired, as the model is supposed to generalize to unseen subjects.

\begin{figure}[tb]
\begin{center}
\centerline{\includegraphics[width=0.9\columnwidth]{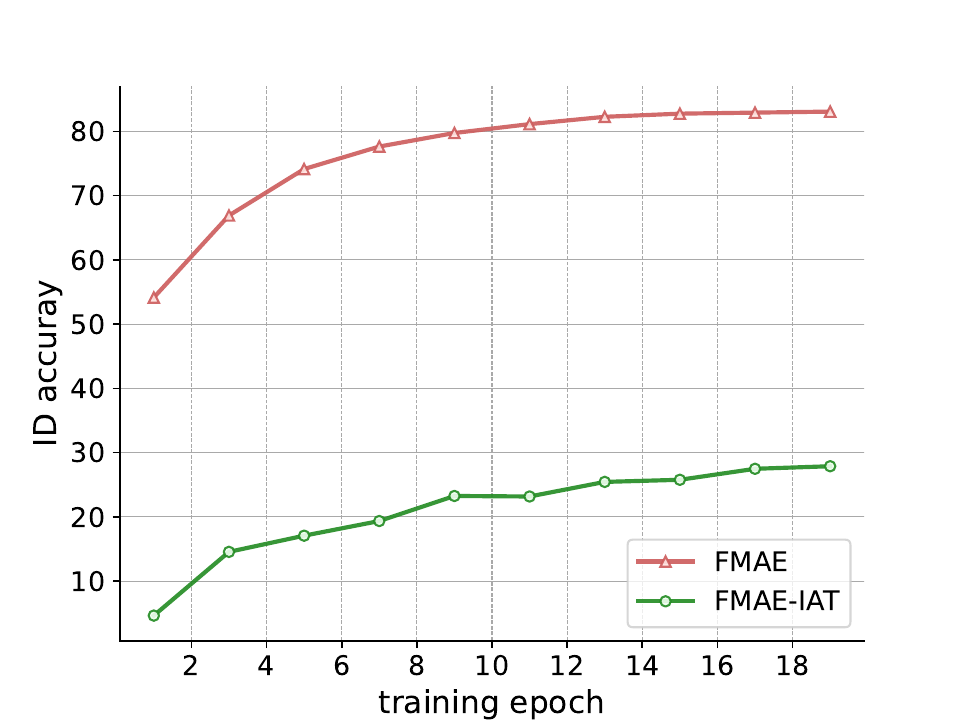}}
\caption{Identity recognition accuracy (\%) evaluated by linear probing on the BP4D dataset. IAT greatly reduces the identity-related features learned by the network backbone $G_{f}(\cdot; \theta_{f})$.}
\label{fig: ID linear prob}
\end{center}
\end{figure}

To demonstrate identity learning in AU detection, we quantitatively and qualitatively evaluate the identity features via linear probing \cite{chen2020simple} and t-SNE \cite{van2008visualizing} technique, respectively. In detail, we apply linear probing on a trained AU detection model (FMAE) and evaluate the identity recognition accuracy on the BP4D dataset, which contains 41 subjects with the identity labels. Specifically, we freeze the backbone $G_{f}(\cdot; \theta_{f})$ of a well-trained FMAE and add a learnable linear classifier on top of the backbone to predict the identity label. For each subject in BP4D, we randomly draw 70 samples for training and 30 samples for testing. The resultant accuracy under linear probing is shown in Figure \ref{fig: ID linear prob}, the red line indicates that FMAE can recognize more than half of people among the 41 subjects even though the model is only trained for one epoch. Given enough training, the identity recognition accuracy can be as high as 83\%. By contrast, IAT significantly alleviates this identity learning issue with 4.6\% accuracy after one epoch of training and 27.9\% accuracy at epoch 19. An interesting phenomenon is that even under the strong identity unlearning regularization, FMAE-IAT seems still to partially learn the identity-based features, by showing 27.9\% accuracy (higher than the random guess accuracy 2.4\%). We believe that the inherent high correlation between training and testing images for each subject provides the possibility for the model to infer the identity by looking at the non-face area.

\begin{figure}[tb]
    \centering
    \begin{subfigure}[b]{0.45\textwidth} 
        \centering        \includegraphics[width=\textwidth]{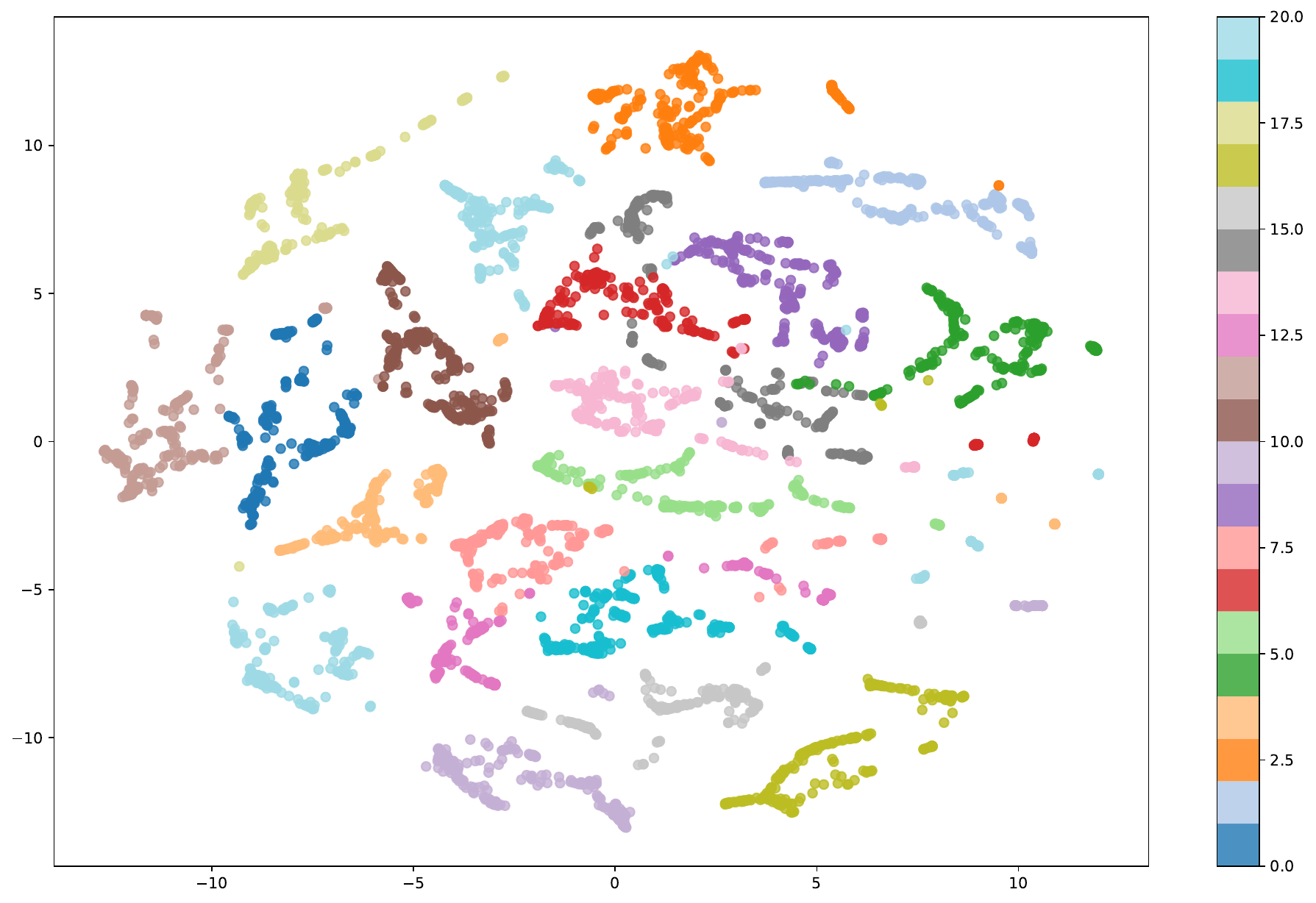}
        \caption{FMAE}
        \label{fig: FMAE tsne}
    \end{subfigure}
    \vspace{0.1cm}
    
    \begin{subfigure}[b]{0.45\textwidth} 
        \centering
        \includegraphics[width=\textwidth]{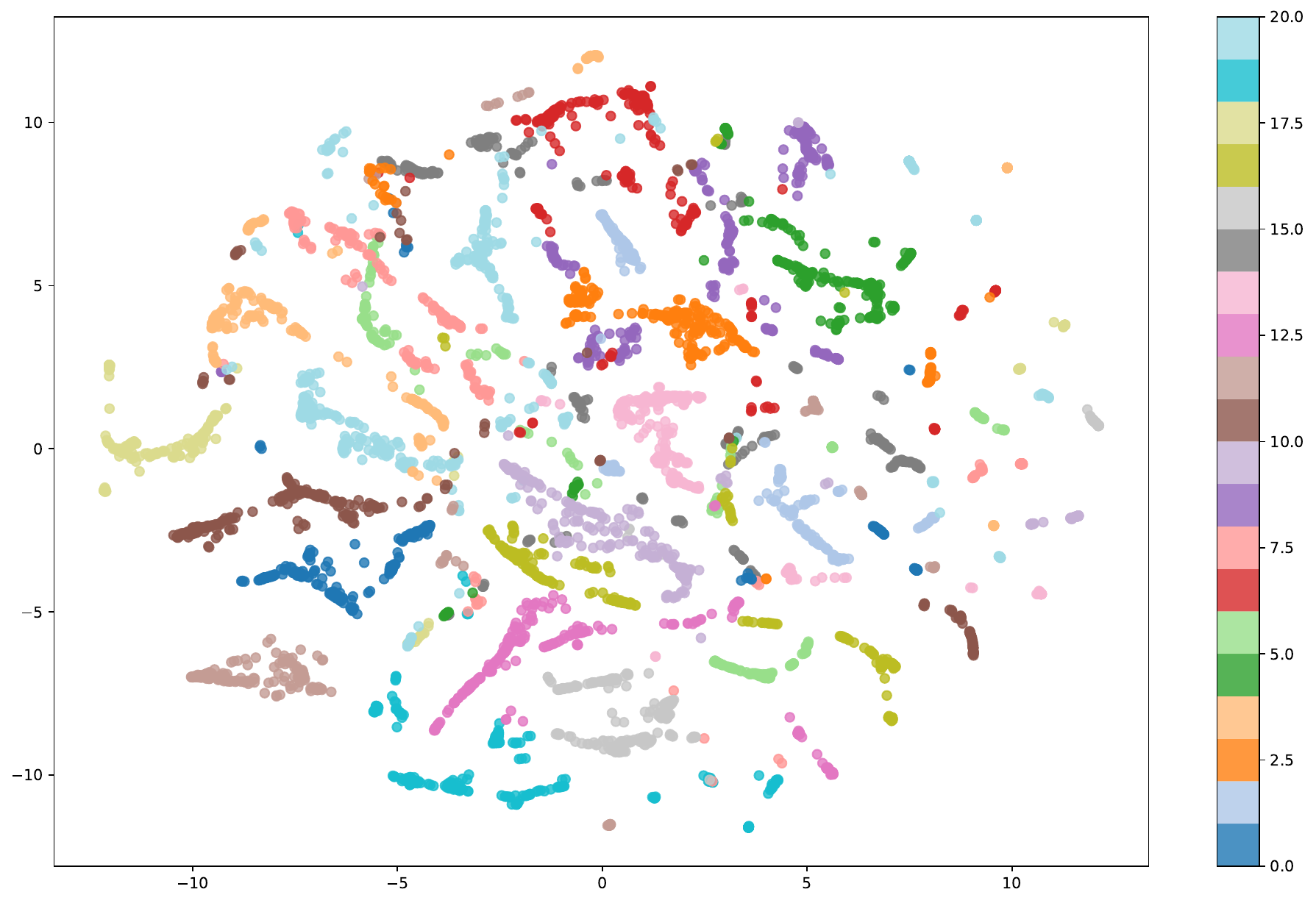}
        \caption{FMAE-IAT}
        \label{fig: FMAE-IAT tsne}
    \end{subfigure}
    \caption{t-SNE visualization of the backbone features on
BP4D dataset regarding the identity labels, each color stands for a subject. Only 20 subjects are visualized for readability even though BP4D contains 41 subjects. FMAE features are more identity-clustered than FMAE-IAT features}
    \label{fig: tsne}
\end{figure}

We also visualize the feature output from the backbone of FMAE and FMAE-IAT using t-SNE and see how these features are clustered according to the identity label. Figure \ref{fig: tsne} presents the t-SNE results for 20 subjects in BP4D (41 subjects in total), given trained FMAE and FMAE-IAT models. It is clear that the identity-based feature clusters in FMAE become less linearly distinguishable (the ID head is a linear layer) under the effect of IAT.

\subsection{IAT mitigates identity shortcut learning}
\label{sec: learning dynamics}
\begin{figure}[tb]
\begin{center}
\centerline{\includegraphics[width=0.9\columnwidth]{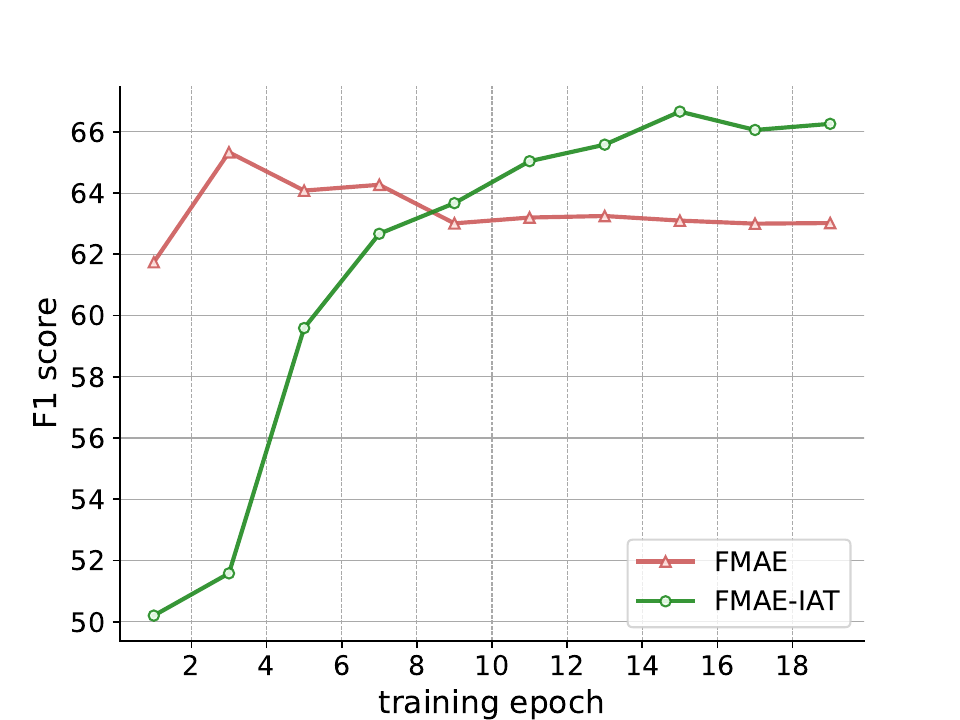}}
\caption{F1 dynamics of FMAE and FMAE-IAT on BP4D+ during training. Fold-2 of BP4D+ is used for visualization.}
\label{fig: F1 learning dynamics}
\end{center}
\end{figure}

After showing that a regular AU model (FMAE) learns the subject identity, we now illustrate that the identity shortcut learning leads to a trivial AU prediction solution that is inferior to the solution delivered by IAT. Concretely, we observe that FMAE and FMAE-IAT have totally different learning dynamics in terms of AU predictions (indicated by F1 score). Figure \ref{fig: F1 learning dynamics} shows the F1 score of both models along the training epochs, where the two models share the same learning rate, batch size and initial training states. It is clear from Figure \ref{fig: F1 learning dynamics} that FMAE is optimized quickly and converges at the third epoch with an F1 score of 65.45\% under the identity shortcut. In contrast, FMAE-IAT learns the AU decision boundary progressively and converges only at epoch 15 with an F1 score of 66.66\%. One can infer that IAT explicitly pushes the backbone $G_{f}(\cdot; \theta_{f})$ away from the identity-related solution region and delivers a better solution for AU detection tasks.

\subsection{Large \texorpdfstring{$||-\lambda \frac{\partial L_{id}}{\partial \theta_{f}}||$} \\ is necessary}
\label{sec: strong adversarial}
In Section \ref{sec: IAT method}, we have mentioned the key design space of IAT: a linear projection layer for the ID head and a large $\lambda$ for the gradient reverse layer. These two factors together ensure the large magnitude of $-\lambda \frac{\partial L_{id}}{\partial \theta_{f}}$ during the adversarial training of the backbone $G_{f}(\cdot; \theta_{f})$. We elaborate here on the specifics of the IAT design space. We postulate that learning the subject identity is relatively easy, since there are many facial components and non-facial cues that can be used for identity recognition. Therefore, a strong regularization of IAT (i.e., a large $||-\lambda \frac{\partial L_{id}}{\partial \theta_{f}}||$) is required to counteract the identity-related learning tendency.

We first show the effect of using different $\lambda$ on the fold-2 of the BP4D dataset. All models share the same training settings except for $\lambda$. In Table \ref{tab: lambda effect}, `Epoch' indicates the training convergence point in terms of the F1 score and $\lambda=0$ represents the group without IAT. We see that a small $\lambda$ ($\lambda=0.02$), such as the one used in \cite{zhang2018identity}, has little gain of F1 score, whereas the large $\lambda$ ($\lambda=1, 2, 3$) yields significant improvement of AU prediction. Moreover, the larger $\lambda$ we use, the more training epochs are required to reach a better optimization point, which is consistent with the phenomenon in Figure \ref{fig: F1 learning dynamics}. Additionally, we perform the ablation study on $\lambda$ using the AU datasets to demonstrate that $\lambda$ is an easy hyper-parameter to tune in practice. Table \ref{tab: lambda ablation} shows that $\lambda$ values within the set of [0.5, 1, 2], which are used across all AU datasets in this paper, consistently result in an improvement in the F1 score.

\begin{table}[tb]
\small
\caption{The effect of different $\lambda$ on BP4D. F1 is reported on fold-2 of BP4D and Epoch means the convergence epoch during training.}
\begin{tabular}{@{}l|lllll@{}}
\toprule
$\lambda$ & 0 & \begin{tabular}[c]{@{}l@{}}0.02\\ (used in \cite{zhang2018identity})\end{tabular} & 1 & 2 & 3 \\ \midrule
F1 & 68.33 & 68.60 & 69.26 & \textbf{69.57} & 69.47 \\
Epoch & 2 & 10 & 20 & 21 & 27 \\ \bottomrule
\end{tabular}
\label{tab: lambda effect}
\end{table}

\begin{table}[tb]
\small
\caption{Ablation study of $\lambda$ on BP4D, BP4D+ and DISFA. F1 scores are reported for one fold of each dataset, with the numbers in parentheses indicating the absolute improvement in F1 compared to the baseline $\lambda=0$.}
\begin{tabular}{@{}lllll@{}}
\toprule
\multirow{2}{*}{Dataset} & \multicolumn{4}{c}{$\lambda$}                                                       \\ \cmidrule(lr){2-5} 
                         & 0     & 0.5                    & 1                      & 2                      \\ \midrule
BP4D                     & 68.81 & 69.22 (+0.41)          & 69.26 (+0.45)          & \textbf{69.57 (+0.76)} \\
BP4D+                    & 65.45 & 66.58 (+1.13)          & \textbf{66.66 (+1.21)} & 66.62 (+1.17)          \\
DISFA                    & 71.06 & \textbf{73.48 (+2.42)} & 73.23 (+2.17)          & 73.01 (+1.95)          \\ \bottomrule
\end{tabular}

\label{tab: lambda ablation}
\end{table}

Furthermore, we show that recognizing identity is a trivial task since we find that a non-linear ID head $G_{f}(\cdot; \theta_{id})$ can still recognize the subjects given the identity-invariant features (regularized by IAT). To investigate this in more detail, we increase the model capacity of the ID head $G_{f}(\cdot; \theta_{id})$ given the backbone trained with a large $\lambda$, and measure the identity loss. Table \ref{tab: mlp effect} shows the results of using different MLP layers for FMAE-IAT under the same regularization strength ($\lambda=2$). The ID loss in Table \ref{tab: mlp effect} suggests that the model gradually learns the identity given some model capacity. By contrast, using the 1-layer MLP (linear projection layer) for the ID head leads to a large ID loss $L_{id}$, thus ensuring the large magnitude of $-\lambda \frac{\partial L_{id}}{\partial \theta_{f}}$. Therefore the linear projection layer is another necessity for IAT in AU detection. The convergence epoch and F1 score in Table \ref{tab: mlp effect} also imply that the 2-layer MLP and 3-layer MLP both converge fast and learn a sub-optimal solution to the AU tasks, which is consistent with our previous observations.   

\begin{table}[tb]
\small
\caption{The effect of different MLPs for the ID head. Epoch in the table shows the convergence epoch during training and ID loss indicates the average identity loss at the convergence epoch using the training set. A higher ID loss implies a lower ID accuracy.}
\begin{tabular}{@{}l|lll@{}}
\toprule
ID head & 1-layer MLP & \begin{tabular}[c]{@{}l@{}}2-layers MLP\\ (used in \cite{zhang2018identity})\end{tabular} & 3-layers MLP \\ \midrule
F1 & \textbf{69.57} & 69.00 & 68.90 \\
ID loss & 0.152 & 0.096 & 0.085 \\
Epoch & 21 & 7 & 6 \\ \bottomrule
\end{tabular}
\label{tab: mlp effect}
\end{table}

\section{Conclusion}
\label{sec: conclusion}
In conclusion, we have proposed to use a masked autoencoder (FMAE) with diverse pre-training for the AU detection task in this paper. We have leveraged a vast and diverse dataset (Face9M) for pretraining, combined with masked image modeling to significantly improve AU detection performance. Moreover, we have demonstrated the identity learning issue and its harmful effect on AU prediction models. The use of Identity Adversarial Training helped in mitigating the model's learning of identity-related features. Also, we elucidated the two design factors of IAT, and our experiments consistently demonstrated superior performance over previous methods, achieving new SOTA results on AU benchmarks like BP4D, BP4D+ and DISFA.

We also noticed that the scaling effect of FMAE pretrained on Face9M has not converged even using the ViT-large model. The potential of using ViT-huge and distilling it into a smaller model for practical use is promising, and we leave this for future work.

{\small
\bibliographystyle{ieee_fullname}
\bibliography{egbib}
}

\end{document}